# Machine Learning to Promote Translational Research: Predicting Patent and Clinical Trial Inclusion in Dementia Research

## Authors

Matilda Beinat*, Julian Beinat, Mohammed Shoaib, Jorge Gomez Magenti*
King's College London, United Kingdom. Alzheimer's Research UK, Cambridge, United Kingdom.

## Correspondence

Matilda Beinat
Strand, London, United Kingdom WC2R 2LS
mailto: matilda.beinat@kcl.ac.uk

Jorge Gomez Magenti
3 Riverside, Granta Park, Cambridge CB21 6AD
jorge.magenti@alzheimersresearchuk.org

# Abstract

Projected to impact 1.6 million people in the UK by 2040 and costing £25 billion annually, dementia presents a growing challenge to society. This study, a pioneering effort to predict the translational potential of dementia research using machine learning, hopes to address the slow translation of fundamental discoveries into practical applications despite dementia's significant societal and economic impact.

We used the Dimensions database to extract data from 43,091 UK dementia research publications between the years 1990-2023, specifically metadata (authors, publication year etc.), concepts mentioned in the paper, and the paper abstract. To prepare the data for machine learning we applied methods such as one hot encoding and/or word embeddings. We trained a CatBoost Classifier to predict if a publication will be cited in a future patent or clinical trial.

We trained several model variations. The model combining metadata, concept, and abstract embeddings yielded the highest performance: for patent predictions, an Area Under the Receiver Operating Characteristic Curve (AUROC) of 0.84 and 77.17% accuracy; for clinical trial predictions, an AUROC of 0.81 and 75.11% accuracy.

The results demonstrate that integrating machine learning within current research methodologies can uncover overlooked publications, expediting the identification of promising research and potentially transforming dementia research by predicting real-world impact and guiding translational strategies.

# Introduction

Almost 1 million people currently live with dementia in the UK,[1] and a recent study forecasts that 1.6 million people will be affected in the UK by 2040.[2] The increasing incidence rate in the 2010s marks a concerning trend reversal compared to other Western nations.[3],[4] The economic burden of dementia-related diseases, like Alzheimer's Disease, is substantial, costing UK society approximately £25 billion annually.[5]

At the same time, the last decade has seen a notable surge in dementia research publications from UK institutions, attributed to increased government investment since the mid-2010s and the rise of organisations like Alzheimer's Research UK.[6] In 2022 alone, over 3,000 dementia research papers were published in the UK.[7]

This burgeoning research landscape reveals a significant gap between scientific advancements and tangible health outcomes,[8],[9] underscoring the urgency to expedite and facilitate the translation of fundamental research into real-world benefits for dementia patients and their families. The vast data processing capabilities of machine learning (ML), analysing extensive data pools within seconds, positions it as a promising potential tool to improve decision-making when addressing this translational challenge.[10]

## The slow pace of progress

The amyloid hypothesis was postulated more than 30 years ago.[11] Only in the past 2 years, after hundreds of unsuccessful clinical trials, the first two drugs targeting beta-amyloid accumulation in the brain demonstrated clinical efficacy in people with early onset Alzheimer's Disease.[12],[13] This contrasts with the cumulative private expenditure on clinical trials (phases 1-4) for Alzheimer's Disease, which was estimated to be $42.5 billion over a 25-year span.[14] This estimate does not include the cost of the research leading up to those phase 1 trials, which is also notorious for being a lengthy and complicated process: the beginning of the "valley of death" in drug discovery.[15]

This is not a problem unique to dementia research. In fact, most biomedical research does not translate into meaningful benefits for the people whom the research is trying to serve,[16] and when it does, it often takes more than two decades to come to fruition.[17]

Explanations about the slow pace of progress in dementia research and neuroscience more generally tend to focus on factors such as the complexity of the human brain,[18],[19] the challenges with reproducibility of preclinical data,[20] or the lower investment in the field of neuroscience compared to other disease areas with similar economic cost to society.[21],[22] While these are all contributing factors, in this manuscript we build on a concept upstream of all them: it is notoriously difficult to identify innovative ideas,[23],[24],[25] and therefore, the scientific community needs new tools to help us decide what pieces of research are more or less likely to lead to translation and patient impact. This is where machine learning, with its capacity to handle vast data pools and to locate trends within the data, can play a significant role.

## Peer-review and innovation

Peer-review processes, crucial in guiding research funding and publication decisions, often consider scientific quality and potential impact. However, it has been found to be an inconsistent process,[26],[27] and the literature is rich in examples of disparities derived from the peer-review process, related to gender,[28],[29] ethnicity,[30] research institution,[31] and geographic location.[32] Two recent examples in the field of dementia research highlighted that the current review model that informs which ideas receive funding and the type of research that is published favours particular researchers over others.[33],[34] This is a considerably well-established phenomenon, to the point that lottery systems to award research funding have started being introduced by funding organisations.[35],[36]

Frameworks to foster translational impact, such as the Framework to Assess the Impact from Translational health research (FAIT),[37] have been developed but primarily focus on post-hoc evaluation of research impact. Such frameworks, while valuable, do not inherently expedite the process of translating discoveries into tangible benefits for patients.

## Novel methodologies

In recent years, innovative methodologies have emerged to enhance the identification and evaluation of translational research. For instance, Manjunath et al. undertook an inventive approach by mapping patent citations to research articles.[38] They revealed that papers cited by patents possess distinct textual, semantic, and author demographic characteristics, which can be indicative of their real-world biomedical impact. Similarly, Nelson et al. delved into an extensive analysis of 43.3 million papers, employing deep learning models that amalgamate metadata and abstract text.[39] Their study concluded that these advanced models are markedly superior to traditional citation metrics in predicting a research paper's inclusion in patents, guidelines, and policies. Complementing these findings, Cao et al. leveraged text mining and predictive modelling to explore how the inherent properties and network positions of scientific concepts facilitate their transition from basic research to practical applications.[40] This approach illuminates the intricate pathway of research evolution from theory to practice.

More recently, Li et al. introduced a novel Translational Progression (TP) measure, rooted in biomedical knowledge representation.[41] By analysing over 30 million PubMed articles, their method dynamically tracks the trajectory of biomedical research along the translational continuum, and they argue that their measure could be used by policymakers to monitor biomedical research with high translational potential in real-time.

Collectively, these methodologies display a significant shift in the landscape of research tools and models. They can prove instrumental in bridging the gap between scientific discovery and practical healthcare applications, a particularly poignant challenge in the field of dementia research.

## Contribution to the field

In this work, we developed an ML model capable of scanning research paper metadata - such as author names, number of authors, research organisations, funders, etc. - and giving a prediction on whether that paper will be cited by patents or clinical trials. This tool could, therefore, facilitate the process of identifying fundamental research with the highest "translational potential".

Alongside paper metadata, we also investigated the impact of content-based features to the models predictive analysis. A model that can help understand the translational potential of research based on the scientific content of publications could have significant implications in the decision-making processes of funders, research organisations or scientific journals.

To the best of our knowledge, this is the first analysis in the dementia research field, using a novel ML framework, with the potential to predict translation of fundamental research.

# Methods

## Data availability

This paper uses bibliometric available through Dimensions API (https://www.dimensions.ai/products/all-products/dimensions-api/). The code used to extract data from Dimensions, the clean data to create the ML model, and the Python code (version 3.12) created to train the CatBoost ML model are available here: https://github.com/MatildaBeinat/ML-for-Translational-Research. Any additional information can be found upon contact with the authors.

## Data

The dataset was downloaded from Dimensions, a scientific database that includes data on 141 million publications, 158 million patents, 7 million grants and 811 thousand clinical trials (as of January 2024). Dimensions consists of a comprehensive list of publications linked via citations allowing for a succinct view of the landscape of today's research, mapping the research cycle from input to output.[7] Metadata on 43,091 publications in the field of dementia research was extracted with python code, using Google Colab as our IDE. These publications were produced by institutions based in the United Kingdom between 1990 and 2023.

The full feature list can be found in Table S1 along with their definition in Table S2. The feature list comprised of:

- categories representing the publications research and disease categorisation,
- health research authority category,
- research activity codes,
- reference id count,
- first author id,
- first author name,
- first author affiliation id,
- first author affiliation country,
- first author affiliation name,
- author count,
- funders,
- journal id and title,
- accessibility,
- organisation country names,
- organisation names,
- concepts,
- and abstracts.

Time sensitive metrics provided by Dimensions such as 'recent citations', 'altmetric score', and 'times cited' were not included as features in the model due to suspicion of label leakage. Dimensions-provided concepts were included based on their importance score to prevent irrelevant concepts from cluttering the ML model. To further make sure that less significant concepts were not included in the feature list we omitted concepts that appeared fewer than twenty times in the dataset. Abstract embeddings of individual publications were included in the feature list [42]. Titles were excluded to avoid duplication, as we found that title words were often repeated in the abstract text. We utilised OpenAI's text embedding ADA-02 tool to process abstracts. Long abstracts had to be truncated to meet the tool limitations. We employed a parallel processing technique for the ADA-02 API calls, to reduce the time it takes to embed all abstracts.[43]

When all features in Table S1 were included in the training set, the model performed best, seen by the model metrics.

Patent and clinical trial data for each publication was accessed and added. This particular data was used to create the label for the ML model.

Data, such as concepts, category research activity code (rac), and category research, condition, and disease categorization (rcdc), were formatted as lists and were one hot encoded. Concepts, category rcdc, category rac, and abstracts required feature reduction in the form of Truncated Singular Value Decomposition (TSVD). Data was extracted and analysed from the database with Dimensions Search Language (DSL, an SQL inspired language) and Python 3.12. Using Python on Google Colaboratory, we organised and selected the data for the ML model. GPT4 - via the interface of chatGPT - was used as an aid to the coding process to optimise, refactor or improve code.[44]

## Machine Learning model

We selected CatBoostClassifier, among the many ML tools available,[45] a modern and popular ML classifier used for its high accuracy and ability to handle numerical and categorical data simultaneously.[46] We used Google Colab as our Python IDE to clean and prepare data, train the model, and review the results. We used Google colab forms for a user-friendly interface, allowing for a simple way to adjust and alter parameters. We used a variety of libraries to support the analysis, from data processing to training and visualisation.

The label was created based on whether the publication was cited in a patent or clinical trial, allowing the model to predict a publications inclusion/exclusion from patent/trial. With a classification model the label required to be binary: 0 or 1. We chose to label a publication with '0' if no prior citations within a patent or clinical trial were found in the database, and '1' for publications with a minimum of 1 citation in a patent or clinical trial. For the purposes of this work, no difference was made in the classification between papers with a single patent citation and papers with multiple patent citations.

We split the data between 75:25 for training and testing respectively. We optimised the model for accuracy, however, the user can choose the performance metric for the model, selecting from Accuracy, Precision, Recall, or F1 Score. Furthermore, the user has the option to select the specific label for training the model, either 'Patents' (label_patents) or 'Clinical Trials' (label_trials)"

We rebalanced our dataset due to a disproportionate number of publications without a citation in patents or clinical trials. The majority of the negative class was randomly down sampled by 85% or 95% for patents and clinical trials respectively. Rebalancing of a majority class label is useful to prevent the model from learning only from the majority class, and thus failing to generalise. Oversampling of the minority class of the dataset was not recommended for our dataset as it has been seen to perform poorly when dealing with large datasets, such as ours.[47] Rebalancing via downsampling led to 1:1.04 balance of 1s and 0s (2647 Ones and 2733 Zeros) in the label for patents and 1:1.54 for clinical trials (660 Ones and 1011 Zeros) (Figure S1). The rebalanced data set was held throughout training and testing of the model.

## Models per label

For each label, label patents and label clinical trials, three models were trained to examine the effect of different feature sets. Model 1, the simplest one for each of the two labels, was trained using publications' metadata only. This model provides a baseline to test the model's performance when including content-based information in the feature set. For both prediction problems, Model 2 was trained using metadata alongside concept embeddings. Finally, Model 3 was trained with metadata, concept and abstract embeddings. The final model consists of the most elaborate thematic data, allowing for the model to learn which characteristics of a publication, including its scientific content extracted from embeddings, led to patent or clinical trial inclusion or exclusion.

The performance for each model was measured by the accuracy, lift score, AUC and precision-recall curves. We defined the lift score as the accuracy of the model minus the prevalence of zeros in the label input, a way to measure the effectiveness of the predictive model. AUC scores were visualised in an ROC curve, which measures the classifiers ability to distinguish between the labels. We also used precision-recall curves, which are more resistant to the bias of an imbalanced datasets.

We further analysed the models capability to predict accurately throughout time by inferring the label of 10 randomly selected publications for each year within our dataset, with equally split positive and negative labels. We plotted the Δ label (real label - predicted label) for each publication with a threshold of 0.5 and -0.5 to visualise which publication was labelled inaccurately by the model.

# Results

## Distribution of publications

Over the period of January 1990 to October 2023, 43,091 dementia research papers were published in the United Kingdom. Of these research papers, 3,026 research papers were cited in a patent and 850 research papers were cited in a clinical trial. This timeframe was selected in order to obtain a sufficiently large number of publications for analysis, and to allow enough time for citations from patents and clinical trials to develop. Of these publications, 36,850 were suitable for analysis, the rest had insufficient data in the features list and were excluded. Taking into account papers published up until 2017, 20,864 papers were used to train and test the model. This selection was made during the data preparation phase (Fig. S1).

Upon exploring our data set, we found that research papers that were cited in a patent were on average cited by 245.93 other papers, while research papers that were not cited in a patent were on average cited by 30.85 other papers. A two-sample unequal variance t-test confirmed the significant difference on citations (p<0.001).

Alongside that, research papers that were cited in clinical trials were on average cited by 382.24 other papers, while papers that were not cited in clinical trials were on average cited by 39.19 other papers. A two-sample unequal variance t-test confirmed the significant difference on citations (p<0.001).

The average time delay from paper publication to patent citation was 4.28 years (Fig. S2). The average time delay from publication to clinical trial citation was 6.57 years (Fig. S3), showing that it takes longer for fundamental research to influence clinical trials than patents. For the training and testing of our model, we chose to exclude papers published after the year 2017, to allow enough time for the papers to have been cited in a patent or clinical trial.

## Model performance

### Label patents

Model 1 returned an AUC of 0.82, with an accuracy of 75.53%, precision of 73.42%, recall of 79.67%, F1 of 74.42%, and lift score of 25.28%. The full model metrics alongside the confusion matrix can be seen in Table 1. The precision-recall curve is shown in Fig 1A.

In contrast, the combination of metadata and concept embeddings as feature input (Model 2) returned an AUC score of 0.83, an accuracy score of 76.28%, precision of 74.44%, recall of 79.67%, F1 of 76.97%, and a lift score of 26.02%. Model 2, trained on both metadata and concepts, outperforms Model 1, which was trained just with metadata as feature input. For a full visualisation of the model metrics refer to Table 2. The precision-recall curve is shown in Fig 1B.

Finally, the most complex model for patent prediction was trained on metadata, concept embeddings and abstract embeddings as feature input (Model 3). This model resulted in the best model of the three trained and tested, returning an AUC score of 0.84, an accuracy value of 77.17%, precision, recall and F1 of 75.21%, 80.72%, and 77.87% respectively with a lift score of 26.91%. The full metrics are found in Table 3, precision and recall curve seen in Fig 1C.

We plotted the three models' AUCs onto one Receiver Operating Characteristic (ROC) curve for a better visualisation of their predictive performance. As seen in Fig 1D, the model that outperforms the rest is model 3 with standard metadata, concept and abstract embeddings as feature input, while the model with only standard metadata as feature input underperforms each model in label trials.

**Table 1**
*Model 1. Metadata model label patents metric values and confusion matrix.*

| Metric | Values | Confusion matrix | | |
|---|---|---|---|---|
| Accuracy | 75.53% | | Predicted (0) | Predicted (1) |
| Precision | 73.42% | True (0) | 483 | 193 |
| Recall | 79.67% | | | |
| F1 | 76.42% | True (1) | 136 | 533 |
| Lift | 25.28% | | | |

**Table 2**
*Model 2. Metadata and concepts model label patents metric values and confusion matrix.*

| Metric | Values | Confusion matrix | | |
|---|---|---|---|---|
| Accuracy | 76.28% | | Predicted (0) | Predicted (1) |
| Precision | 74.44% | True (0) | 493 | 183 |
| Recall | 79.67% | | | |
| F1 | 76.97% | True (1) | 136 | 533 |
| Lift | 26.02% | | | |

**Table 3**
*Model 3. Metadata, concepts and abstracts model label patents metric values and confusion matrix.*

| Metric | Values | Confusion matrix | | |
|---|---|---|---|---|
| Accuracy | 77.17% | | Predicted (0) | Predicted (1) |
| Precision | 75.21% | True (0) | 498 | 178 |
| Recall | 80.72% | | | |
| F1 | 77.87% | True (1) | 129 | 540 |
| Lift | 26.91% | | | |

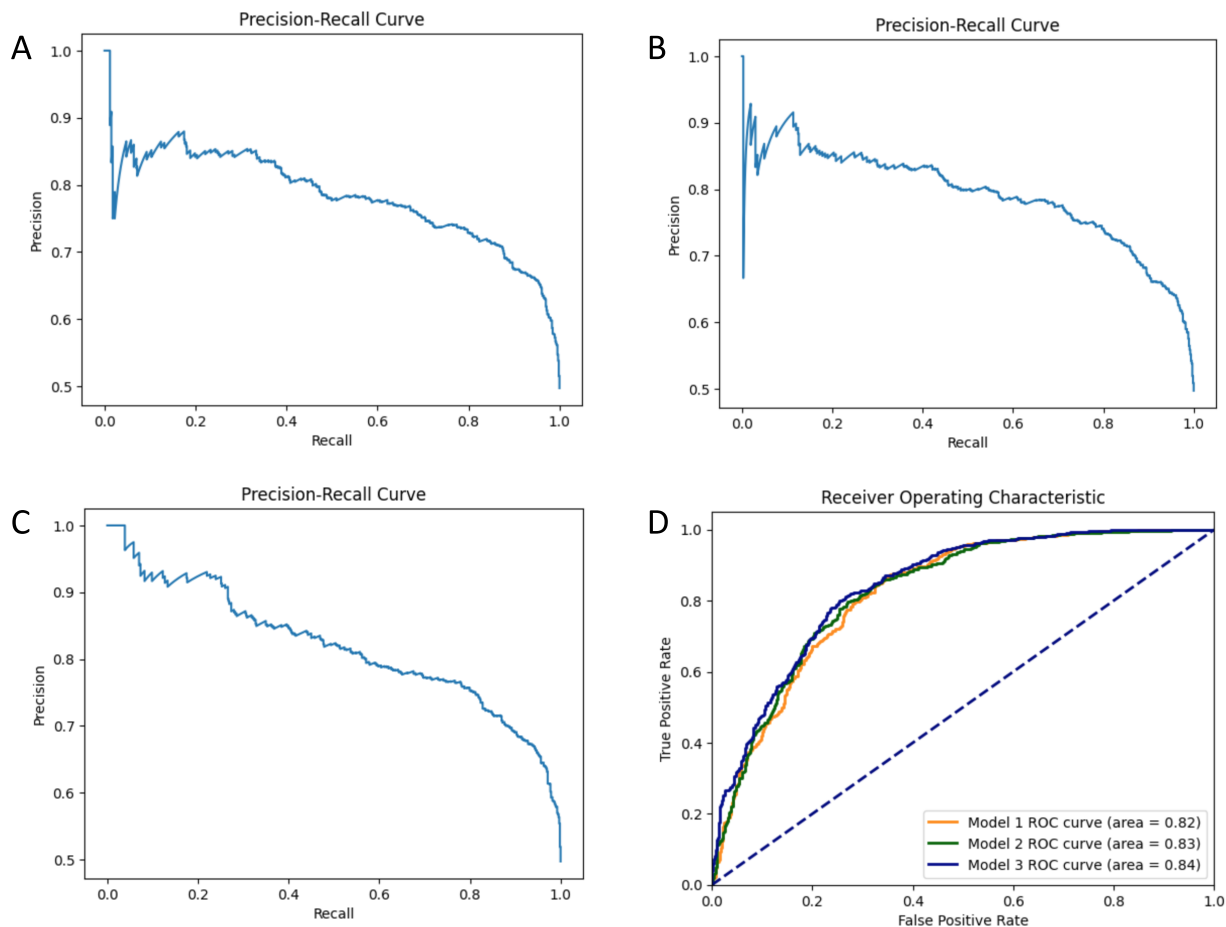

**Fig 1. Predictive performance depicted by precision-recall curves and AUROC for label patents.**
(A-C) Precision-recall curves for model 1, 2 and 3 respectively. Model 1 trained and tested on only metadata, model 2 trained on metadata and concept embeddings, model 3 trained on the aforementioned including abstract embeddings for prediction of label patents. Models are plotted against their respective precision and recall outputs. Further shown are the ROC curves (D) where the yellow curve depicts model 1, green for the model trained on a combination of metadata and concept embeddings, and model 3 in blue. Each model seen here is trained and tested for the years 1990-2017 with a random data split of 75:25 for training and testing respectively.

Label trials

Model 1 resulted with an AUC of 0.77, accuracy of 73.44%, precision of 70.15%, recall of 56.97%, F1 of 62.88%, and lift score of 12.92%. This model achieved the lowest metric output out of each model for both translational outcomes. A full layout of the metrics for model 1 label clinical trials can be seen in [Table 4](), precision-recall curve in [Fig 2A]().

Model 2 resulted with an AUC score of 0.78, accuracy of 73.68%, precision of 69.50%, recall of 59.39%, F1 of 64.05%, and lift score of 13.16%. Model 2 for clinical trials performed similarly to model 1, however has a slightly higher accuracy and lift score, full metrics and confusion matrix in [Table 5](). The precision-recall curve is seen in [Fig 2B]().

Our final model, the model that outperformed the previous 2 models for label clinical trials, was trained and tested on a combination of metadata, concept embeddings and abstract embeddings as feature input. This model outputted an AUC score of 0.81, accuracy of 75.11%, precision of 72.26%, recall of 60.00%, F1 of 65.56%, and lift score of 14.59%. The model metrics can be seen in [Table 6](), [Fig 2C]() for precision-recall curve.

To better visualise how each model for label trials performed, we plotted each AUC score onto one ROC curve ([Fig 2D]()). The ROC curve in this case depicts how each model performed compared to one another, the highest performing model being the final model with metadata, concept and abstract embeddings as feature input, and the lowest performing model with the lowest accuracy and AUC being the initial model with only standard metadata as feature input.

**Table 4**
*Model 1. Metadata model label trials metric values and confusion matrix.*

| Metric | Values | | Confusion matrix | |
|---|---|---|---|---|
| Accuracy | 73.44% | | Predicted (0) | Predicted (1) |
| Precision | 70.15% | True (0) | 213 | 40 |
| Recall | 56.97% | | | |
| F1 | 62.88% | True (1) | 71 | 94 |
| Lift | 12.92% | | | |

**Table 5**
*Model 2. Metadata and concepts model label trials metric values and confusion matrix.*

| Metric | Values | | Confusion matrix | |
|---|---|---|---|---|
| Accuracy | 73.68% | | Predicted (0) | Predicted (1) |
| Precision | 69.50% | True (0) | 210 | 43 |
| Recall | 59.39% | | | |
| F1 | 64.05% | True (1) | 67 | 98 |
| Lift | 13.16% | | | |

**Table 6**
*Model 3. Metadata, concepts and abstracts model label trials metric values and confusion matrix.*

| Metric | Values | | Confusion matrix | |
|---|---|---|---|---|
| Accuracy | 75.11% | | Predicted (0) | Predicted (1) |
| Precision | 72.26% | True (0) | 215 | 38 |
| Recall | 60.00% | | | |
| F1 | 65.56% | True (1) | 66 | 99 |
| Lift | 14.59% | | | |

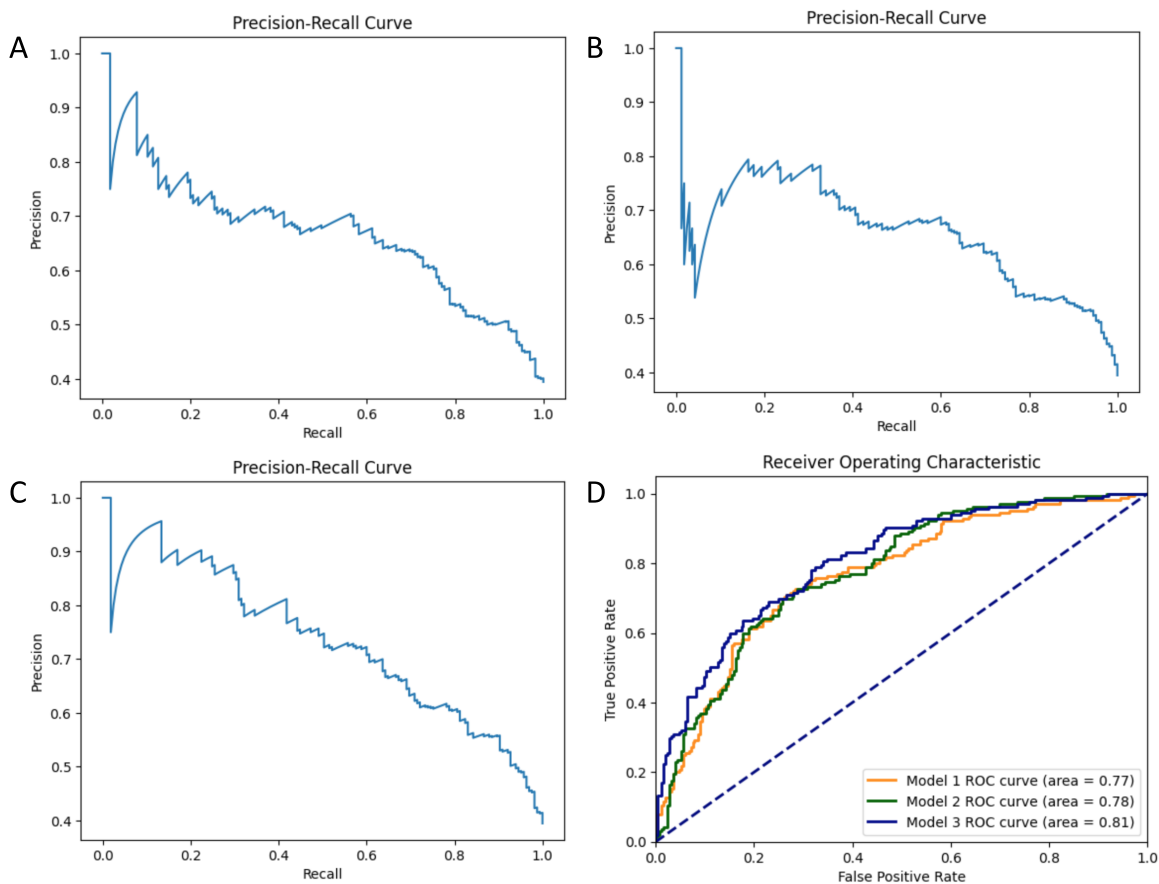

**Fig 2. Predictive performance depicted by precision-recall curves and AUROC for label trials.**
(A-C) Precision-recall curves for model 1, 2 and 3 respectively. Model 1 trained and tested on only metadata, model 2 trained on metadata and concept embeddings, model 3 trained on the aforementioned including abstract embeddings for prediction of label clinical trials. Models are plotted against their respective precision and recall outputs. Further shown are the ROC curves (D) where the yellow curve depicts model 1, green for model 2, and model 3 seen in blue. Each model seen here is trained and tested for the years 1990-2017 with a random data split of 75:25 for training and testing respectively.

## Application to other fields

In order to test the models applicability to other fields, we used the same methodology applied to cancer research papers. We experimented on the prediction label for patent inclusion with publications for breast cancer published between the years 2010 and 2015. We included standard metadata, concepts and abstract embeddings as feature input, downsampling the majority label by 85%. The model resulted with an AUC of 0.81, accuracy of 75.27%, precision of 71.54%, recall of 89.44%, F1 of 75.73%, and lift score of 23.24%. The confusion matrix and full metric outputs are in Table S3, alongside the ROC curve and precision-recall curve for the breast cancer model in Fig. S4.

Predicted label patents

The data we are analysing is highly time sensitive. This stems from the inherent nature of publications, which require time to be noticed, cited, and to contribute value to the field – a process beyond our control. Consequently, the label might be affected by time-dependent factors, for which we do not have explicit controls in the models, yet we ratified the problem as best we could by removing highly time dependent features (e.g. recent citations, times cited, altmetric scores). To understand the time sensitivity of our models we tested whether the model is capable of predicting the label consistently through time using our test data set. We explored this by randomly selecting 5 publications with label 0 and 5 with label 1 for every year within our timeframe 1990 - 2017, analysing 270 publications in total. This was done to see the performance of the model through three different decades. We compared the real label of each publication with the predicted label calculated with model 3 for label patents (metadata, concepts and abstract embeddings included as features for prediction). The predicted label in this case is the percent chance the model expects the publication to be cited in a patent: below 50% the model will assign a label 0, above 50% the model assigns a label 1.

To determine the difference between the predicted label and the real label, we plotted the Δ label for each individual publication. For example: a publication with label 1 and predicted label 0.75, the Δ label is 0.25. Therefore the closer the Δ label value is to 0, the better the model performed (Fig. 3).
We calculated the predictive values to depict the degree of inaccuracy of the model. We found that the model performs similarly through time (74% accuracy in the 90s, 73% accuracy in the 2000s, and 79% accuracy in the 2010s) (Table S4).

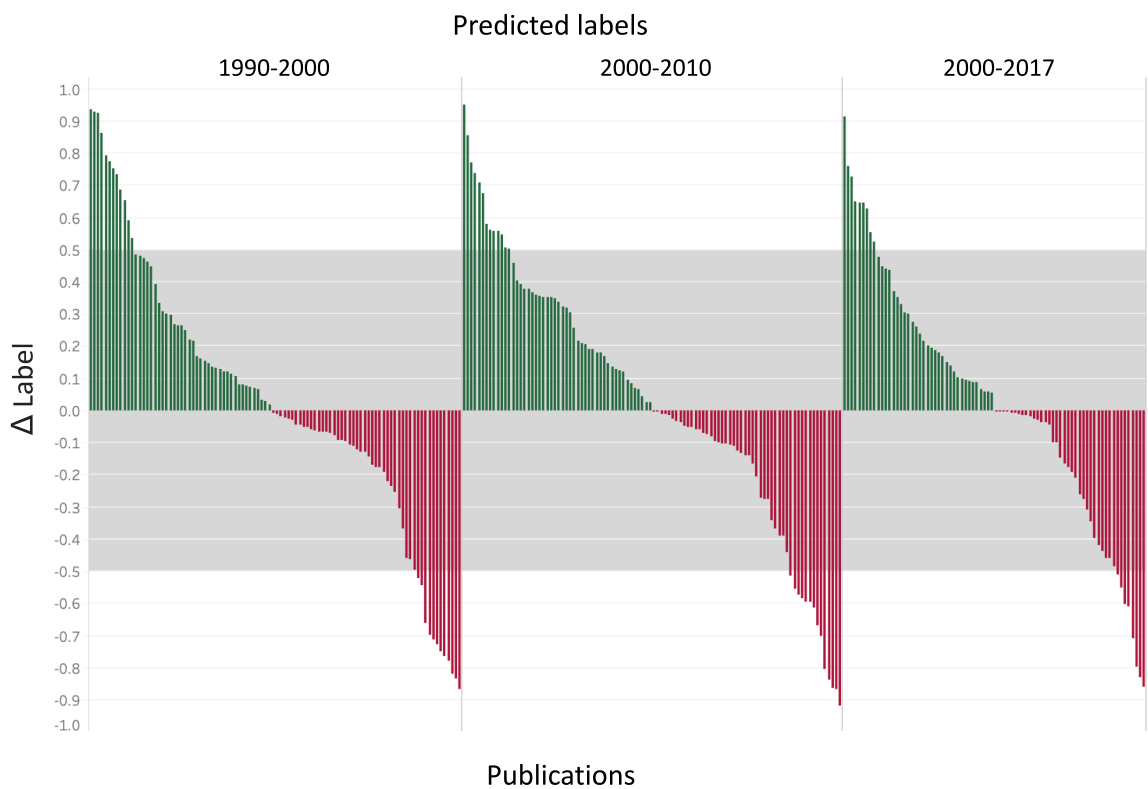

**Fig. 3. Predicted label versus actual label using model 3 label patents.**
Label 1 is shown in green and label 0 is shown in red. Δ label is plotted for 10 publications per year for the timeframe 1990-2017. The grey box depicts the threshold that decides whether the publication will be cited in a patent or not. The predicted labels within the grey box are the correct predictions, all other publications have been mislabeled by the model.

# Discussion

This study represents, to our knowledge, the first study to predict the translational potential of dementia research using machine learning. By leveraging a comprehensive dataset from Dimensions, encompassing 43,091 publications, and related data on patents, and clinical trial citations, our ML model provides a novel approach to help highlight research papers that will lead to a patent or clinical trial inclusion. This work addresses a critical need in the field of dementia research, where the translation of fundamental research into practical applications has been notably slow, despite the substantial societal and economic burden of the disease. The integration of ML in this context demonstrates a promising avenue for addressing the translational gap, which research funders and organisations have not been able to successfully close.

It is within this context that we see the most promise for models like this one: as a tool that can help aid in the decision-making process. Within our data pool of 43,091 publications, only 3,026 publications - or 8.99% of the data pool - were cited in a patent or clinical trial. In this work, we present a model that can accurately predict whether a paper will be cited by a patent 77.17% of the time and can identify 80.72% of the papers that will go on to be cited by patents. We believe funders, research organisations and even scientific publishers could benefit from incorporating tools like this one into their decision-making processes.

## Model interpretation

The classification models we created can be used in conjunction with current translational research methods, such as the FAIT method.[37] By integrating our ML model with these methods, we can uncover overlooked publications and expedite the identification of promising research. This synergy is crucial for advancing the field of dementia research, where the slow pace of progress and high failure rates in clinical trials have been long standing challenges.

Classification models are most useful to categorise particular datasets into predefined classes. Such models have one of two outputs: probabilistic or categorical. Our model outputs two possible categories: 1 or 0. The application of ML in this context is particularly promising, as it enables the handling of vast and complex datasets, a task impractical for traditional methods, simply by categorisation. Our approach, which combines various data features, demonstrates the capacity of ML models to unravel intricate patterns and associations within academic research that has the capability to predict real-world impact. Working with large data sets, such as this, it is crucial to work with models that are effective and can achieve high accuracy levels, such as classification models. Further, classification models can handle very unbalanced datasets, and with the nature of our dataset, it is more likely to encounter unbalanced datasets, with preference towards negative labels. Most importantly, they are very suitable for real-time predictions, essential in the field of translational research where predicting a paper's potential for further applications can speed up the process from basic research to clinical practice.

In developing our model, we drew inspiration from pioneering works in the field: Manjunath et al correlate patent citations with research articles which revealed distinct textual, semantic, and author demographic characteristics that could predict real-world impact.[38] This methodology informed our decision to use patent citations as a measure of a paper's translational potential. Nelson et al on the other hand utilised deep learning models to analyse extensive databases, they demonstrated the superiority of advanced ML models over traditional citation metrics in predicting a research paper's inclusion in patents and policies.[39] We incorporated similar techniques, emphasising the role of abstract embeddings in predicting research impact. Cao et al further leveraged text mining and predictive modelling, exploring the transition of scientific concepts from basic research to practical applications.[40] Their approach influenced our analysis of the content and context of research papers, specifically the inclusion of concepts and themes identified in publications. Finally, Li et al. introduced a novel Translational Progression (TP) measure, tracking the trajectory of biomedical research along the translational continuum.[41] Our methodology was inspired by their dynamic approach, using biomedical knowledge representation to identify high-potential research.

Our approach demonstrates the capacity of ML to analyse complex data sets. The inclusion of concept and abstract embeddings substantially improves the ML model's predictive accuracy, indicating the importance of thematic and lexical analysis in classifying publications. Importantly, our model can predict the potential of newly published papers for inclusion in patents or clinical trials, demonstrating its applicability in real-time assessment of research impact.

## Model performance

We show that through analysis of standard metadata points, such as the number of authors in a publication, prediction outputs are inferior to those attained via high dimensional vectors. Our analysis shows that the inclusion of concept and abstract embeddings, depicting overarching themes and trends within the publication, substantially improves the ML models prediction accuracy. The inclusion of concept and abstract embeddings shows us that the ML model extracts important information about themes necessary to classify publications.
Importantly, the ML model is capable of predicting whether newly published papers have potential for inclusion in patent or clinical trials. This signifies that through the correct frameworks it is feasible to execute high fidelity outputs that provide the field with a succinct path for translational output.

Contrary to traditional approaches that heavily rely on citation metrics, our findings suggest that these are not the sole indicators of a publication's impact. Our ML model effectively identifies research likely to lead to patent or clinical trial citations without depending on time-sensitive data such as altmetric scores or the number of citations. This finding is pivotal in redefining the metrics used for assessing the translational potential of research.
We now have the option to incorporate machine learning models, with the capability to analyse complex thematic based data, with our quantitative metrics, enabling us to identify themes and trajectories with the highest translational potential. We can use a supervised learning method, such as a classification model, alongside unsupervised dimensionality reduction methods that capture large texts, translating from a high dimensional vector to a low dimensional space, to enrich our data analyses. Research suggests that using an entire body of text, rather than concepts and abstract embeddings only, may result in higher fidelity.[39] On the other hand, there has been opposing research stating that while embeddings may be effective in shorter texts, model performance has been seen to increase with abstract embeddings, challenging the notion that whole text embeddings may be superior.[48] However, the exploration of whole text embeddings versus abstract embedding only was outside the scope of our research.

## Limitations

While our study marks significant progress, it is not without limitations. The largest challenge lies with the nature of the publications, particularly their requirement of time for patent or clinical trial inclusions. We attempted to rectify this limitation by removing time sensitive features that could leak the label to the model. This in turn allows for the application of the model to newly published papers and removes time as an obstacle in the models prediction.

The secondary challenge lies in the reliance on citations in patents and clinical trials as proxies for translational impact. This approach, while practical, may not fully capture the multifaceted nature of research impact. However, based on research by Manjunath et al., papers cited by patents seem to have distinct characteristics that are indicative of real-world impact. Based on the models increasing performance when including thematic based features, we believe our model is extracting intricate characteristics for patent and clinical trial prediction.[38] We believe that due to the models increased predictive analysis, with features such as concept and abstract embeddings, provides a deeper understanding of otherwise impossible to understand information.[39] Even so, patent-cited articles are identified as a unique subset of biomedical literature, characterised by differences in scientific domain, research team composition, and language use. This uniqueness poses a challenge in generalising findings across different subsets of literature. This constraint however is amended by the models applicability to multiple different domains, seen by its application to breast cancer research by our model alongside research presented by Cao et al. and Li et al.[40],[41]

The complexity of the classification model, or any ML model, is a double-edged sword. On one hand, it is capable of identifying and learning from complex patterns in data samples. On the other hand, this very complexity makes it challenging to pinpoint which features of the model are most influential in determining its performance. Essentially, it's difficult to simplify the model to just its most impactful elements due to its intricate nature. This implication prevents the understanding of the model. However, this limitation is not the real problem to translational research, in reality it is crucial and the most beneficial method. It is important to understand the difference between simplicity of the method and explainability. While current frameworks for translational research are explainable and easily understood by the wider public, it is clear that their function is not effective. ML models, whether supervised or unsupervised, can be complex to explain. Although it's often difficult to isolate individual features to understand their specific contributions, these models function in a straightforward and effective way. They provide a more accurate representation of the complex research they are used to analyse.[43] The simplicity of models can be disadvantageous as they do not adequately represent the complexity of research such as that encountered here.

Our model's feature weighting is not entirely transparent, and it might be subject to biases.[33], [34], [35], [36] Biases may be the result of multiple factors, such as unbalanced data for certain classes of papers, or underrepresentation of certain subpopulations in scientific literature. While there are several methods to test models against specific biases,[49] this test was beyond the scope of the current research. At the same time, it should be noted that our model goes beyond simple citation counts used by current methodologies, analysing content and themes, potentially offering a more balanced view of a paper's merit.

## Concluding paragraph

Machine learning models that use standard metadata and complex thematic analysis (such as concepts and abstract embeddings), are able to distinguish papers with a high potential for translation. The application of classification models alongside current methodologies, such as the FAIT method, has the potential to create high fidelity outputs, providing a trajectory for plausible translational applications. In our case, this approach can be applied to predict the potential of a publications inclusion in a patent or clinical trial, providing a pathway towards impact. Our work advocates for a shift towards the use of novel methodologies, such as machine learning models. These applications are more crucial than ever, especially with the decreasing productivity in translational research seen over the past 50 years, urgently requiring action.

## Funding

M.B. was supported by King's College London to conduct this work as part of her MSc's dissertation. J.G.M and M.S. were employed by Alzheimer's Research UK at the time of this work.

## Author Contribution

Conceptualisation: M.B. and J.G.M; methodology:, M.B., J.G.M., and J.B.; code: J.B. and M.B.; data analysis: M.B., J.G.M., and J.B.; data visualisation: M.B. and J.B.; original draft writing, M.B. and J.G.M.; draft editing and reviewing, M.B., J.B., J.G.M., D.P., E.B. and M.S.

## Acknowledgements

Our heartfelt thanks go to Euro Beinat and Dafydd Prole for their invaluable time and support in reviewing and providing feedback on the manuscript. Finally, we would like to thank Alzheimer's Research UK for hosting M.B. as an intern between the months of May and August 2023.

# Supplementary data

**Table S1**
*Full feature list with metadata that is found within the Dimensions database, alongside features included in each model. The definition of each feature can be found in the supplementary data ([Table S2](#)).*

| Feature | Categorical Yes/No | Features included for label patents and clinical trials | | |
|---|---|---|---|---|
| | | Model 1 | Model 2 | Model 3 |
| Category rcdc (research, condition, and disease categorization) | Y | ✓ | ✓ | ✓ |
| Category hra (health research authority) | Y | ✓ | ✓ | ✓ |
| Category hrcs (health research classification system) rac (research activity code) | Y | ✓ | ✓ | ✓ |
| Reference ids count | N | ✓ | ✓ | ✓ |
| Recent citations | N | | | |
| Altmetric | N | | | |
| Relative citation ratio | N | | | |
| Times cited | N | | | |
| First author id | N | ✓ | ✓ | ✓ |
| First author name | Y | ✓ | ✓ | ✓ |
| First author affiliation id | N | ✓ | ✓ | ✓ |
| First author affiliation country | Y | ✓ | ✓ | ✓ |
| First author affiliation name | Y | ✓ | ✓ | ✓ |
| Funder countries | Y | ✓ | ✓ | ✓ |
| Authors count | N | ✓ | ✓ | ✓ |
| Funders | Y | ✓ | ✓ | ✓ |
| Journal id | N | ✓ | ✓ | ✓ |
| Journal title | Y | ✓ | ✓ | ✓ |
| Open access | Y | ✓ | ✓ | ✓ |

| Research organisation country names count | N | ✓ | ✓ | ✓ |
| Research organisation names | Y | ✓ | ✓ | ✓ |
| Research organisation names count | N | ✓ | ✓ | ✓ |
| Research organisation country names | Y | ✓ | ✓ | ✓ |
| Concepts | Y | | ✓ | ✓ |
| Abstracts | Y | | | ✓ |

**Table S2**
*Feature definitions of features considered for the ML models.*

| Feature | Definition |
| --- | --- |
| Category hra | Categorised by health research areas. |
| Category hrcs rac | Categorised by health research classification system and their research activity codes. |
| Category rcdc | Categorised by research, condition, and disease. |
| Reference ids | There are IDs for publications in the reference list of the original publication. |
| Recent citations (excluded from feature list) | The number of citations the publication received in the last 2 years. |
| Altmetric (excluded from feature list) | The attention score based on the weight count of attention received by the publication across the internet. |
| Relative citation ratio (excluded from feature list) | Citation performance of the paper relative to other papers in the same field. |
| Times cited (excluded from feature list) | The number of times the publication has been cited. |
| First author id | The ID of the first author that created the publication |
| First author name | The name of the first author that created the publication. |
| First author affiliation id | The affiliation ID of the first author in the publication |
| First author affiliation country | The affiliated country of the first author in the publication |
| First author affiliation name | The name of the affiliation of the first author in the publication |
| Authors count | The number of authors in the publication. |
| Funder countries | The country of the organisations that fund the publication. |
| Funders | Organisation funding the publication. |
| Journal id | The journal that the publication belongs to. |
| Journal title | The title of the journal that the publication belongs to |
| Open access | Publications category determining whether it is free access: Gold (free), Bronze (available on publisher's website), Hybrid (available under open licence), Green (available in open access repository), Closed (not free). |
| Research organisation names | The names of the organisations the authors are associated with. |
| Research organisation names count | The number of research organisations related to the authors in the publication |
| Research organisation country names | The country names of the organisations the authors are associated with. |
| Concepts | Keywords in the publication. |
| Abstracts | Entire abstract of individual publications |

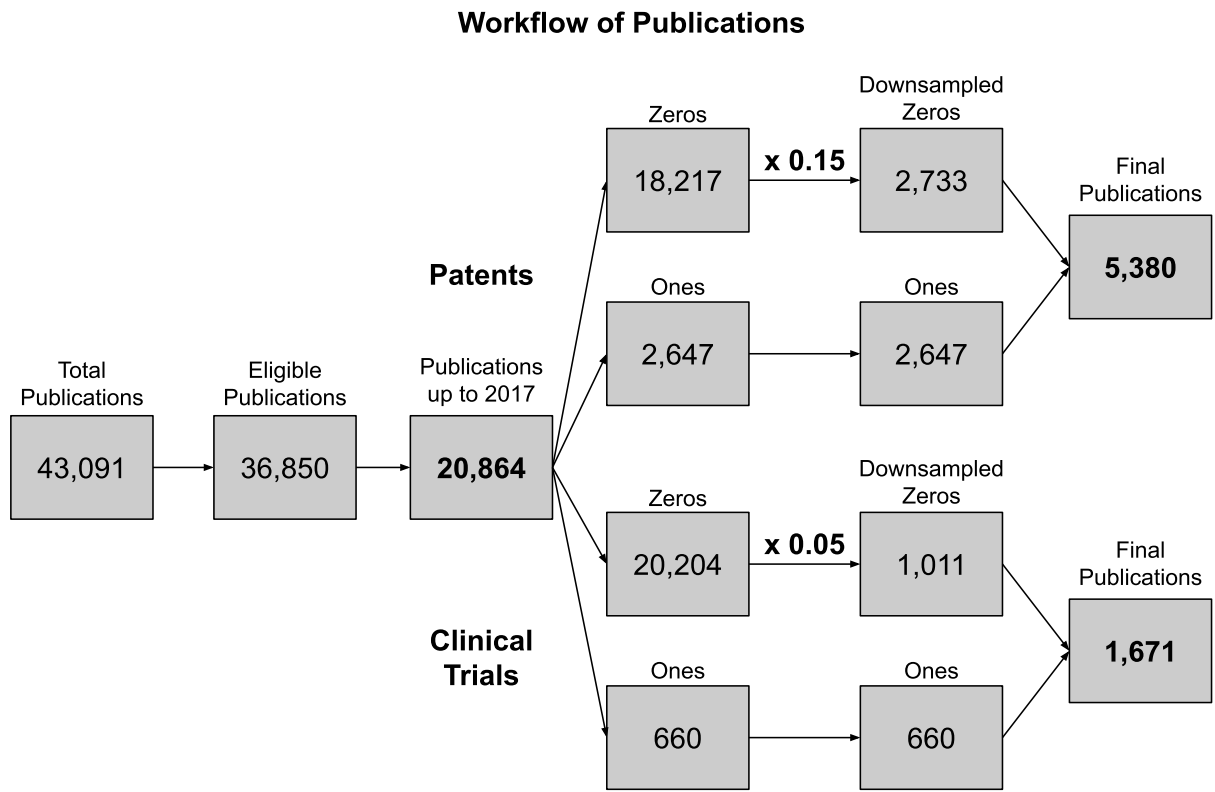

**Fig. S1. Workflow diagram of publications.**
Workflow description to achieve final publications for both patents and clinical trials.

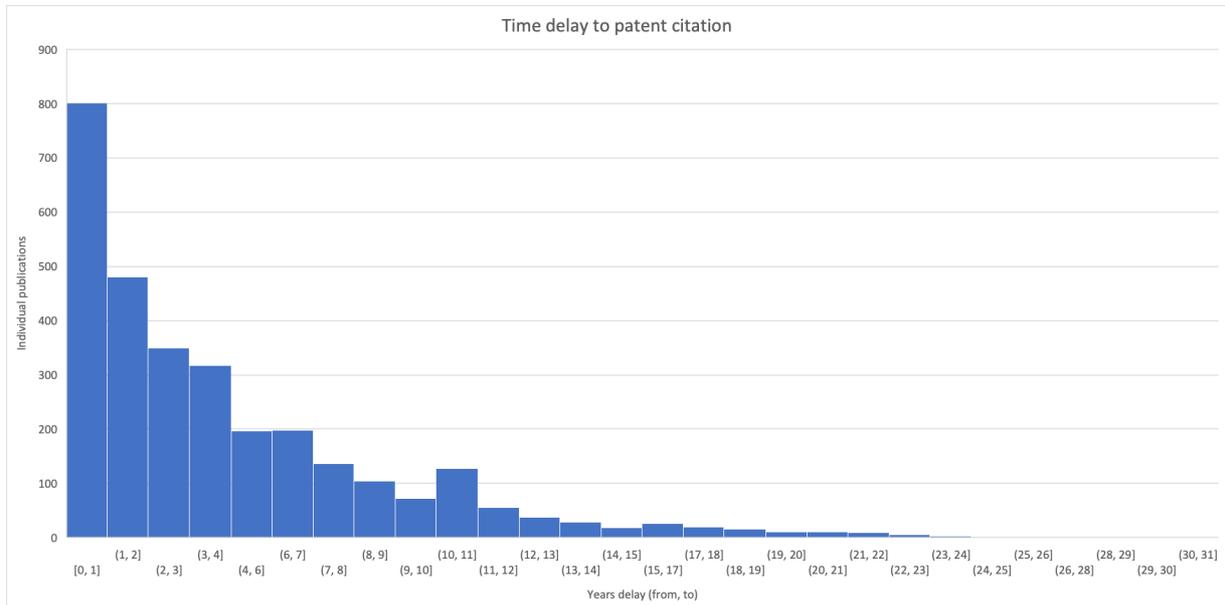

**Fig. S2. Time delay from paper publication to initial patent citation.**
The average time delay from paper publication to initial patent citation is of 4.28 years, with a standard deviation of 4.57.

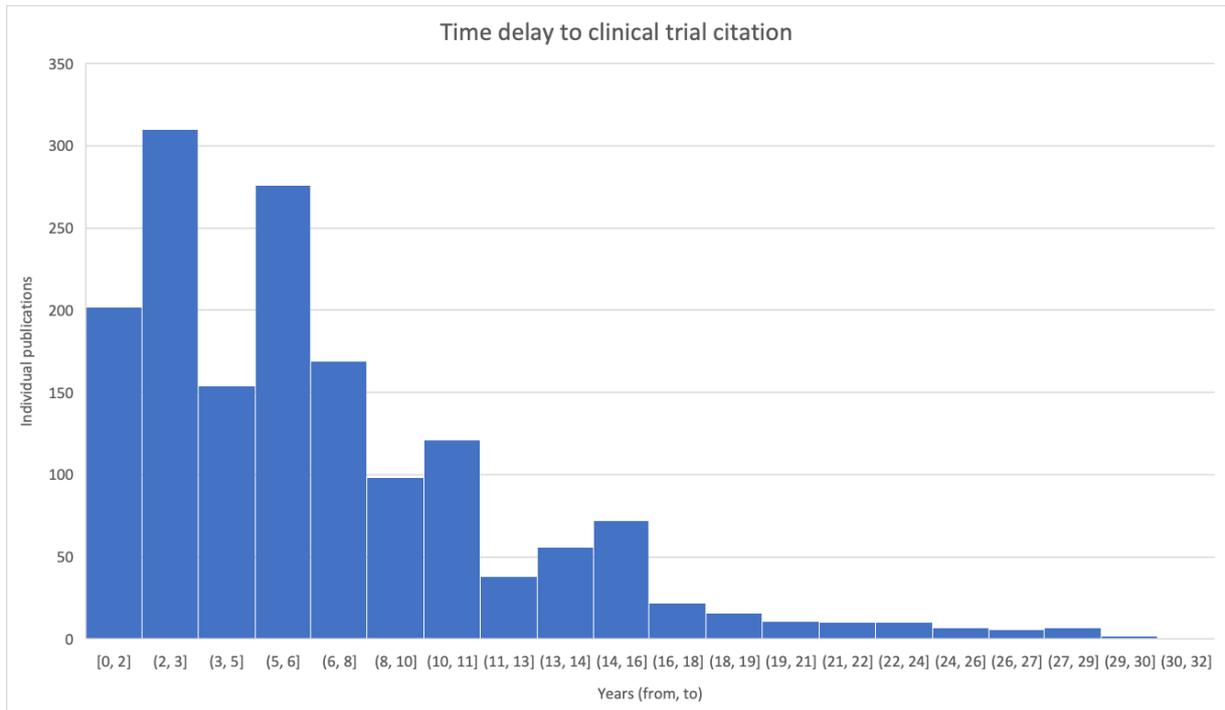

**Fig. S3. Time delay from paper publication to initial clinical trial citation.**
The average time delay from paper publication to clinical trial citation was 6.57 years, standard deviation of 5.77.

**Table S3**

*Breast cancer experiment model for label patents. Metric values and confusion matrix.*

| Metric | Values | Confusion matrix | | |
|---|---|---|---|---|
| Accuracy | 75.27% | | Predicted (0) | Predicted (1) |
| Precision | 71.54% | True (0) | 172 | 72 |
| Recall | 80.44% | | | |
| F1 | 75.73% | True (1) | 44 | 181 |
| Lift | 23.24% | | | |

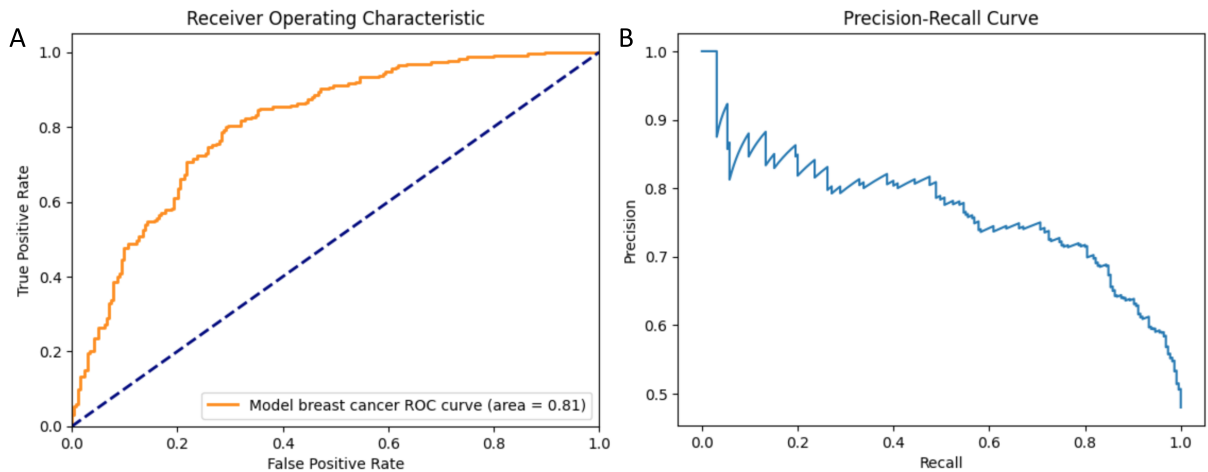

**Fig. S4. ROC and precision-recall curves for breast cancer experiment model label patents.**
Shown is the ROC curve (A) where the yellow curve depicts the model trained on standard metadata, concept and abstract embeddings as feature input. The model is trained and tested for the years 2010-2015 with a random data split of 75:25 for training and testing respectively. (B) Precision-recall curves for the breast cancer experiment model plotted against the respective precision and recall outputs.

**Table S4**
*Confusion matrices for each decade in the predicted label patents (Figure 3).*

|  |  |  | Predicted label | |
|---|---|---|---|---|
|  |  |  | **0** | **1** |
| **Real label** | **1990-2000** | **0** | 38 | 12 |
|  |  | **1** | 12 | 36 |
|  | **2000-2010** | **0** | 36 | 14 |
|  |  | **1** | 13 | 37 |
|  | **2010-2017** | **0** | 32 | 8 |
|  |  | **1** | 9 | 31 |